\documentclass{article}

\usepackage{arxiv}
\usepackage{graphicx}%
\usepackage{multirow}%
\usepackage{amsmath,amssymb,amsfonts}%
\usepackage{amsthm}%
\usepackage{mathrsfs}%
\usepackage[title]{appendix}%
\usepackage{xcolor}%
\usepackage{textcomp}%
\usepackage{manyfoot}%
\usepackage{booktabs}%
\usepackage{algorithm}%
\usepackage{algorithmicx}%
\usepackage{algpseudocode}%
\usepackage{listings}%
\usepackage{mathrsfs}
\usepackage[utf8]{inputenc} 
\usepackage[T1]{fontenc}    
\usepackage{hyperref}       
\usepackage{silence}
\usepackage{url}            
\usepackage{booktabs}       
\usepackage{amsfonts}       
\usepackage{nicefrac}       
\usepackage{microtype}      
\usepackage{lipsum}
\usepackage{graphicx}
\graphicspath{ {./images/} }

\usepackage{amsthm}

\theoremstyle{plain}
\newtheorem{theorem}{Theorem}[section]
\newtheorem{lemma}[theorem]{Lemma}

\theoremstyle{definition}
\newtheorem{definition}[theorem]{Definition}

\theoremstyle{remark}

\makeatletter
\renewcommand{\@toptitlebar}{}
\renewcommand{\@bottomtitlebar}{}
\makeatother

\usepackage{fancyhdr}
\title{Kantorovich-Type Stochastic Neural Network Operators for the Mean-Square Approximation of Certain Second-Order Stochastic Processes}

\author{
  Sachin Saini \qquad Uaday Singh
  \\[1ex]
  Department of Mathematics\\
  Indian Institute of Technology Roorkee\\
  Roorkee, 247667, India
  \\[1ex]
  \href{mailto:sachin_saini@ma.iitr.ac.in}{\texttt{sachin\_saini@ma.iitr.ac.in}},
  \href{mailto:uaday.singh@ma.iitr.ac.in}{\texttt{uaday.singh@ma.iitr.ac.in}}
}

\begin{document}
\maketitle
\begin{abstract}
Artificial neural network operators (ANNOs) have been widely used for approximating deterministic input-output functions; however, their extension to random dynamics remains comparatively unexplored. In this paper, we construct a new class of 
\textbf{Kantorovich-type Stochastic Neural Network Operators (K-SNNOs)} in which randomness is incorporated not at the coefficient level, but through \textbf{stochastic neurons} driven by stochastic integrators. This framework enables the operator to inherit the probabilistic structure of the underlying process, making it suitable for modeling and approximating stochastic signals. We establish mean-square convergence of K-SNNOs to the target stochastic process and derive quantitative error estimates expressing the rate of approximation in terms of the modulus of continuity. Numerical simulations further validate the theoretical 
results by demonstrating accurate reconstruction of sample paths and rapid decay of the mean square error (MSE). Graphical results, including sample-wise approximations and empirical MSE behaviour, illustrate the robustness and effectiveness of the proposed stochastic-neuron-based operator.
\end{abstract}

\keywords{Stochastic neuron, Stochastic process, Artificial neural network operators,  K-SNNOs, Mean square approximation.\\
\textbf{MSC Classification: 41A30, 41A58, 47A58, 60H25, 92B20.}
}

\section{Introduction}

Learning from experience is a fundamental capability of neural networks (NNs) as they interact with their environments. 
The remarkable success of NNs across science and engineering stems from their ability to approximate relationships between inputs and outputs. 
This universal approximation property has been rigorously established for deterministic feed-forward neural networks (FNNs) by Cybenko~\cite{cybenko1989approximation}, Funahashi~\cite{funahashi1989approximate}, and Hornik \textit{et al.}~\cite{hornik1989multilayer}. 
A typical one-hidden-layer FNN can be expressed as
\begin{equation}\label{NN definition}
    \mathbf{N}_n(\mathbf{t})
    =\sum_{j=0}^{n} c_j \,
    \sigma\!\left(\mathbf{a}_j\cdot\mathbf{t}+b_j\right),
\end{equation}
where $\mathbf{t}=(t_1,\dots,t_m)\in\mathbb R^m$, the parameters 
$\mathbf a_j\in\mathbb R^m$ and $c_j,b_j\in\mathbb R$, and $\sigma$ denotes the activation function. 
These classical results provide the mathematical foundation for the modern theory of neural approximation.

\vspace{0.2cm}
During the last decade, classical approximation theory has been substantially extended to deep and structured neural networks. 
Bölcskei \textit{et al.}~\cite{bolcskei2019optimal} proved that deep neural networks (DNNs) achieve the same best-$M$-term approximation rates as affine systems such as wavelets and shearlets, establishing optimal accuracy-complexity trade-offs for sparsely connected architectures. 
More recently, Yang \textit{et al.}~\cite{yang2025rates} investigated convolutional neural networks (CNNs) and established minimax-optimal learning rates for regression and classification. 
These advances firmly position neural networks within the framework of nonlinear operator approximation theory, establishing explicit relationships between architecture, smoothness, dimensionality, and convergence order.

\vspace{0.2cm}
Despite these deterministic results, many real-world environments exhibit inherently stochastic behaviour. 
For instance, financial markets fluctuate due to unpredictable socio-economic factors, climate and weather systems 
contain random atmospheric disturbances, wireless communication signals suffer from noise and interference, 
population dynamics and biological systems evolve with random birth-death events, queueing systems such as hospitals, 
banks and web servers experience random arrival and service patterns, sensor measurements in robotics are corrupted by 
noise, and traffic flow varies due to uncertain driver behaviour. 
These examples highlight that deterministic models alone are insufficient to capture the uncertainty present in such 
settings, thereby motivating the need for stochastic frameworks.

Consequently, there is growing interest in developing theoretical models that enable neural networks to learn random functions or stochastic processes in the mean-square sense. 
Several early contributions to stochastic neural networks (SNNs) include the works of Amari \textit{et al.}~\cite{amari1992information-boltezman}, Conti \textit{et al.}~\cite{conti1994approximation}, and Zhao \textit{et al.}~\cite{zhao1996recurrent-SNN}. 
Belli \textit{et al.}~\cite{belli1999artificialNN} demonstrated that certain classes of SNNs can approximate stochastic processes in in the mean-square sense..
More recently, Fabiani~\cite{fabiani2025random} introduced randomized physics-informed neural networks (RPNNs), proving that randomly generated activations can interpolate arbitrary data sets with probability one. 
In parallel, stochastic neural network architectures have been analyzed through stochastic differential equations and sample-wise backpropagation, providing rigorous foundations for learning under stochastic dynamics~\cite{archibald2024numerical}. 
These developments extend universal approximation theory to networks with random or stochastic activations and motivate the study of neural operators for random processes.

\vspace{0.2cm}
In this context, a neural network activated by stochastic neurons can be defined as
\begin{equation}\label{Stochastic N.N definition}
    \eta_n(t,\omega)=\sum_{j=1}^{n} c_j\,\eta_j(t,\omega),
\end{equation}
where $t\in\mathcal T=[c,d]$ and $\omega\in\Omega$, and the stochastic neurons are given by the random integral activations
\[
\eta_j(t,\omega)
   =\int_{\mathcal T}\sigma(nt-j,ns-j)\,dY_s,
\]
with $\{Y_s(\omega)\}$ denoting a stochastic process with orthogonal increments~\cite{belli1999artificialNN,turchetti2004stochastic}. 
Such constructions are closely related to measure-valued formulations of stochastic particle systems and mean-field limits~\cite{isaacson2022mean}, which provide analytical insight into convergence and stability properties of large stochastic neural ensembles.

\vspace{0.2cm}
Parallel to these developments, the theory of neural network operators (NNOs) offers a constructive approximation framework in which network coefficients, weights, and thresholds are explicitly prescribed. 
Cardaliaguet and Euvrard~\cite{cardaliaguet1992approximation} first introduced bell-shaped and squashing-type NNOs for univariate and multivariate functions. 
Later, Anastassiou~\cite{anastassiou1997rate,anastassiou2000rate} established quantitative convergence results using the modulus of continuity.

These operator-based approaches have since been applied to the mean function (which is a deterministic function) of stochastic processes, such as Brownian motion and time-separating processes. 
Notably, Makovoz~\cite{makovoz1996random-APP.with-NN} and Anastassiou and co-authors~\cite{anastassiou2022brownian-app.byNNO,anastassiou2024brownian-timeseprating,anastassiou2023approximation-TIMEsepatingNNOs-MDPI,anastassiou2023neural-Time-sepratingNNO} demonstrated the effectiveness of NNOs in approximating the mean function of a stochastic process.

In addition, modern operator-structured neural architectures such as SlimTrain~\cite{newman2022slimtrain} have shown that separability, stability, and efficient stochastic approximation can play a central role in operator learning.
Further connections to stochastic neuron models can be found in mean-field analyses of interacting particle systems, such as the threshold-based neural dynamics studied by Inglis and Talay~\cite{inglis2015mean}.
Together, these works highlight the significance of operator-theoretic and stochastic frameworks in neural approximation. A stochastic NNOs can be constructed in two distinct ways: either by introducing randomness into the coefficients of the operators( one can see \cite{saini2025constructive}), or by employing stochastic activation functions (stochastic neurons). 

\vspace{0.2cm}
Motivated by these deterministic and stochastic advances, the present work introduces a new class of \emph{Kantorovich-type Stochastic Neural Network Operators} (K-SNNOs). 
These operators extend deterministic Kantorovich NNOs to stochastic settings by incorporating stochastic neuron activations and stochastic integral structures. 
From a theoretical standpoint, the proposed K-SNNOs preserve positivity, linearity, and stability while achieving mean-square convergence analogous to optimal neural approximation rates~\cite{yang2025rates}. 
From an application perspective, K-SNNOs provide a flexible mathematical framework for modeling and approximating stochastic processes, bridging deterministic operator theory, stochastic analysis, and modern neural approximation theory.
\section{Preliminaries}\label{Section2}
We denote by \( L^2(\Omega, \mathcal{F}, \mathbb{P}) \) the set of all stochastic processes \( X_t(\omega) \) for \( t \in \mathcal{T} = [c,d] \) on a fixed probability space \( (\Omega, \mathcal{F}, \mathbb{P}) \). Additionally,  for each \( t \), the random variable \( X_t(\omega) \), satisfies the following conditions
\[
E[X_t(\omega)] = 0 \quad \text{and} \quad E[|X_t(\omega)|^2] < \infty,
\]

where \( \Omega \) is the set of sample points \( \omega \), \( \mathcal{F} \) is the Borel sigma-algebra, and \( \mathbb{P} \) is the probability measure defined on \( \mathcal{F} \). The symbol \( E(\cdot) \) denotes the expectation and \(X_t(\omega)\) is called second-order stochastic process.

For a given stochastic process $\{Y_s(\omega),s\in \mathcal{T}\}$ with orthogonal increments, i.e., 
$$ E\{(Y_{s_2}-Y_{s_1})(Y_{s_4}-Y_{s_3})\}=0; \quad s_1\leq s_2\leq s_3\leq s_4,$$ 
and for a given arbitrary function $\zeta(t,s)\in L^2(\mathcal{T}\times\mathcal{T})$, the stochastic integral in the mean square sense is defined as 
\begin{equation}
    X_t(\omega)= \int_{\mathcal{T}}{\zeta(t,s)dY_
    s}.
\end{equation}
The key feature of the stochastic integral is its ability to establish an isometric mapping between the spaces \( L^2(\mathcal{T} \times \mathcal{T}) \) and \( L^2(\Omega, \mathcal{F}, \mathbb{P}) \). For details about stochastic integral, one can see \cite{doob1990stochastic}. 

In this paper, we restrict attention to those second-order processes which 
admit a deterministic kernel representation (canonical representation) of the form
\begin{equation}\label{eq:kernel-representation}
    X_t(\omega)=\int_{\mathcal{T}}\zeta(t,s)\,dW_s(\omega),
    \qquad t\in\mathcal{T},
\end{equation}
for some kernel 
\[
    \zeta \in L^2(\mathcal{T}\times\mathcal{T}), \qquad 
    \int_{\mathcal{T}}\int_{\mathcal{T}} |\zeta(t,s)|^2\,ds\,dt <\infty.
\]

\medskip
\noindent\textbf{Representation via a deterministic kernel.}
Let $(W_t)_{t\in\mathcal{T}}$ be a standard Wiener process on a finite
interval $\mathcal{T} $ and let 
$\{X_t(\omega)\}_{t\in\mathcal{T}}$ be a second-order process, i.e.
$X_t(\omega) \in L^2(\Omega, \mathcal{F}, \mathbb{P})$ for all $t\in\mathcal{T}$.

We are interested in representations of the form
\[
X_t(\omega) = \int_{\mathcal{T}} \zeta(t,s)\,dW_s, \qquad t\in\mathcal{T},
\]
for some \emph{deterministic} kernel $\zeta \in L^2(\mathcal{T}\times\mathcal{T})$.

A natural setting where this is possible is the following.

Let $\{X_t(\omega)\}_{t\in\mathcal{T}}$ be a centred Gaussian process with 
covariance function
\[
R(t,u) := \mathbb{E}[X_t(\omega) X_u(\omega)], \qquad t,u \in \mathcal{T}.
\]
Assume that there exists a kernel 
$\zeta \in L^2(\mathcal{T}\times\mathcal{T})$ such that
\begin{equation}\label{eq:cov-factorization}
R(t,u) = \int_{\mathcal{T}} \zeta(t,s)\,\zeta(u,s)\,ds,
\qquad t,u \in \mathcal{T}.
\end{equation}
Then, on a suitable probability space carrying a Wiener process
$(W_t)_{t\in\mathcal{T}}$, there exists a version of $\{X_t(\omega)\}$ such that
\begin{equation}\label{eq:wiener-representation}
X_t(\omega) = \int_{\mathcal{T}} \zeta(t,s)\,dW_s, \qquad t\in\mathcal{T},
\end{equation}
and the process defined in \eqref{eq:wiener-representation} is a centred 
Gaussian process with covariance $R$.

\noindent
In particular, a necessary condition for the existence of a deterministic
kernel representation \eqref{eq:wiener-representation} is that
\[
R(t,u) = \mathbb{E}[X_t(\omega) X_u(\omega)]
\]
admits the factorization \eqref{eq:cov-factorization}. 
This means that the covariance operator of $X_t(\omega)$ is of the form
\[
R(t,u) = \langle \zeta(t,\cdot), \zeta(u,\cdot)\rangle_{L^2(\mathcal{T})}.
\]

Conversely, not every $L^2(\Omega, \mathcal{F}, \mathbb{P})$ process admits a deterministic kernel
representation, this is only possible when the covariance admits the
factorisation \eqref{eq:cov-factorization}. Thus, the class considered
here forms a proper subset of all second-order processes, characterised
by Hilbert-Schmidt (or more generally trace-class) covariance structure.
Despite this restriction, the class remains rich and includes several
important models, such as the Ornstein-Uhlenbeck process, fractional
Gaussian-type processes, and other kernel-driven Gaussian systems.

\begin{definition}\cite{costarelli2024asymptotic}
	A function $\sigma:\mathbb{R} \to \mathbb{R}$ is said to be sigmoidal if it is measurable and satisfies the limiting behaviour:
	\begin{center}
		$\displaystyle \lim_{t \to -\infty} \sigma(t) = 0 \quad \text{and} \quad \lim_{t \to +\infty} \sigma(t) = 1.$
	\end{center}
\end{definition}

Throughout this work, we focus on a non-decreasing sigmoidal function \(\sigma\) that fulfils $\sigma(1)<1$, along with the following additional assumptions.
\begin{enumerate}
    \item[(P1)] The function \(\sigma(t) - \frac{1}{2}\) is odd,
    \item[(P2)] The function \(\sigma\) is twice continuously differentiable  on \(\mathbb{R}\) and concave for \(t \geq 0\),
    \item[(P3)] As \(t\) \(\xrightarrow{}\) \(-\infty\), we have \(\sigma(t) = \mathcal{O}(|t|^{-1 -\gamma})\) for some \(\gamma > 0\).
\end{enumerate}

For the aforementioned sigmoidal function, we define the density functions as
\begin{equation}\nonumber
     \mathcal{L}^\sigma(t)=\frac{1}{2}{\left[\sigma(t+1)-\sigma(t-1)\right]}, \hspace{.5cm}  t \in\mathbb{R},
\end{equation}
and the multivariate density function as 
\begin{center}
    $ \Phi^\sigma(\mathbf{t})=\prod\limits_{i=1}^{m}\mathcal{L}^\sigma(t_i), $ where $\mathbf{t}=(t_1,...,t_m)\in \mathbb{R}^m.$
\end{center}

\begin{definition}{\textbf{Discrete absolute moment of order} $\beta$:}  
	Let $\beta \geq 0$. The function $\mathcal{L}^\sigma$ is said to possess a discrete absolute moment of order $\beta$ if  
	\[ 
	\mathcal{M}_{\beta}(\mathcal{L}^\sigma) = \sup_{t \in \mathbb{R}} \sum_{j \in \mathbb{Z}} \left| \mathcal{L}^\sigma(t - j) \right| \cdot |t - j|^\beta  
	\]  
	is finite.  
\end{definition}

Here, we present the key properties of $\mathcal{L}^\sigma(\cdot)$.
\begin{lemma}\label{lemma-density_fun. properties}
For any $\mathcal{L}^\sigma(\cdot)$ as defined above,
\begin{enumerate}
    \item[(i)] $\mathcal{L}^\sigma(t)\geq 0$ for every $t\in \mathbb{R}$ and $\lim\limits_{t\xrightarrow{}\pm\infty}\mathcal{L}^\sigma(t)=0;$
   
    \item[(ii)] For every $t\in \mathbb{R}, \sum\limits_{j\in \mathbb{Z}}{\mathcal{L}^\sigma(t-j)}=1,\mbox{ and } \left\|\mathcal{L}^\sigma(t)\right\|_1=\int_{\mathbb{R}}{\mathcal{L}^\sigma(t)dt=1;}$ 
    
        \item[(iii)] Let $t\in[c,d]$, $n\in \mathbb{N}$, on the condition that $\lceil nc \rceil\leq \lfloor nd\rfloor-1.$ Then,
       $$ \sum_{j=\lceil nc \rceil }^{\lfloor nd \rfloor -1}{\mathcal{L}^\sigma(nt-j)}\geq \mathcal{L}^\sigma(2)>0; $$
       
    \item[(iv)] $\mathcal{L}^\sigma(t)=\mathcal{O}(|t|^{-1-\gamma})$ as $t\xrightarrow{}\pm\infty;$ 
    \item[(v)] The series $\sum\limits_{j\in\mathbb{Z}}{\mathcal{L}^\sigma(t-j)}$ converges uniformly on any compact subset of $\mathbb{R};$
    \item[(vi)] $ \mathcal{M}_0(\mathcal{L}^\sigma)= \sup\limits_{t\in \mathbb{R}}{\sum\limits_{j\in \mathbb{Z}}|\mathcal{L}^\sigma(t-j)|}<\infty;$ 
   
\end{enumerate}    
\end{lemma}
 For detailed proofs, one can refer to \cite{costarelli2014convergence}.
 
Let us now revisit the single-layer multivariate Kantorovich-NNOs introduced by Costarelli  \cite{costarelli2014convergence}.
\begin{definition}
Let $f: \mathfrak{R}=\prod\limits_{i=1}^{m}[c_i,d_i]\xrightarrow{}\mathbb{R}$ be a locally integrable function, $n \in \mathbb{N}$ such that $\lceil nc_i \rceil \leq \lfloor nd_i \rfloor -1$. The multivariate linear positive Kantorovich-type NNOs-$\mathbf{K}_n(f,\cdot)$, activated by the sigmoidal function $\sigma$ are define as
\begin{equation}\label{eq.22(multi, kantoro)}
\mathbf{K}_n(f,\mathbf{t})= \frac{\sum\limits_{j_1=\lceil nc_1 \rceil}^{\lfloor nd_1 \rfloor-1}... \sum\limits_{j_m=\lceil nc_m \rceil}^{\lfloor nd_m \rfloor-1} \left[n^m\int_{\mathfrak{R}_{\mathbf{j},m}}f(\mathbf{v})d\mathbf{v}\right]\Phi^\sigma(n\mathbf{t-j})}{\sum\limits_{j_1=\lceil nc_1 \rceil}^{\lfloor nd_1 \rfloor-1}...\sum\limits_{j_m=\lceil nc_m \rceil}^{\lfloor nd_m \rfloor-1}\Phi^\sigma(n\mathbf{t-j})},     
\end{equation}
\end{definition}
 where $n\in \mathbb{N}$, $\mathbf{v}=(v_1,v_2,...,v_m) 
     \in \mathfrak{R}_{\mathbf{j},m}= \left[\frac{j_1}{n},\frac{j_1+1}{n}\right]\times...\times\left[ \frac{j_m}{n},\frac{j_m+1}{n}\right] $ and $\mathbf{j}=(j_1,j_2,...,j_m)\in\mathbb{Z}^m.$
     
We will now discuss the pointwise and uniform convergence and quantitative estimates of the above-defined Kantorovich operators studied in \cite{costarelli2022quantitative-esti.Normalized-and-kantoro., costarelli2014convergence}.
\begin{theorem}
	Suppose that \( f: \mathfrak{R} \to \mathbb{R} \) is a bounded function. Then, for every point \( \mathbf{t} \in \mathfrak{R} \) at which \( f \) is continuous, the following limit holds:
	\[
	\lim_{n \to +\infty} \mathbf{K}_n(f, \mathbf{t}) = f(\mathbf{t}).
	\]
\end{theorem}

\begin{theorem}\label{Kn-L^presult}
	Let \( 1 \leq p < +\infty \). Then, for any function \( f \in L^p(\mathfrak{R}) \), the following convergence in \( L^p \)-norm holds:
	\[
	\lim_{n \to +\infty} \left\| \mathbf{K}_n(f, \cdot) - f \right\|_p = 0.
	\]
\end{theorem}

\section{Kantorovich-type stochastic NNOs}
To define KSNNOs, we will first explain a stochastic neuron and demonstrate some of its properties.

\begin{definition}
Consider sigmoidal function $\sigma$ and the two-dimensional density function $ \Phi^\sigma(t,s)=\mathcal{L}^\sigma(t)\mathcal{L}^\sigma(s)$ as defined in section \ref{Section2}. The stochastic integral 
$$ \varphi(t,\omega):= \int_{\mathcal{T}}{\Phi^\sigma(t,s)}dY_s$$ 
is referred to as a stochastic neuron.
\end{definition}
We will now demonstrate some basic properties of the stochastic neurons defined above.    
\begin{lemma}\label{Properties of S.neuron} 
For any sigmoidal function $\sigma$ defined in Section \ref{Section2}, we have
\begin{itemize}
    \item[(i)] $E\left|\varphi(t,\omega)\right|=0; \mbox{ if } E|Y_t|=0.$ 
    \item[(ii)] $E\left|\varphi(nt-j,\omega)\right|^2= \lambda_Y^2 \left(\mathcal{L}^\sigma(nt-j)\right)^2\left\|\mathcal{L}^\sigma(\cdot)\right\|_2,$ where \( \lambda_Y^2 \) is the variance of the increments of \( Y_t(\cdot). \) 
    \item[(iii)] $E\left|\int_{\mathcal{T}}\left(\frac{\Phi^\sigma(nt-j, ns-j)}{\sum\limits_{j=\lceil nc \rceil}^{\lfloor nd\rfloor-1}\Phi^\sigma(nt-j, ns-j)}\right)dY_s \right|^2=\frac{\lambda^2_Y\left\|\mathcal{L}^\sigma(\cdot)\right\|^2_2}{n^2\mathcal{L}^\sigma(2)^4}\left(\mathcal{L}^\sigma(nt-j)\right)^2.$
\end{itemize}
\end{lemma}
\begin{proof}
\begin{itemize}
\item[(i)] 
    Let $\sigma$ be any sigmoidal function as defined in Section \ref{Section2}. Then $\Phi^\sigma(t,s) \in L^2(\mathcal{T}\times\mathcal{T})$ and the stochastic integral $\int_{\mathcal{T}} \Phi^\sigma(t,s) dY_s$ is well-defined in the mean square sense w.r.t. the orthogonal process $Y_s(\omega)$.
    
    Now, if $E|Y_s|=0,$ then
    \begin{align}\nonumber
    E\{|\varphi(t,\omega)|\}&= E\left|\int_{\mathcal{T}}\Phi^\sigma(t,s)dY_s\right|&&\\[5pt] \nonumber
    & =\int_{\mathcal{T}}\mathcal{L}^\sigma(t)\mathcal{L}^\sigma(s)E\left|dY_s\right|&&\\[5pt] \nonumber
    & = {\mathcal{L}^\sigma(t)}\int_{\mathcal{T}}\mathcal{L}^\sigma(s)E\left|dY_s\right|=0.&&\\ \nonumber
\end{align} 
\item[(ii)] We have
\begin{align}\nonumber
   E|\varphi(nt-j,\omega)|^2 &= E\left|\int_{\mathcal{T}}\Phi^\sigma(nt-j,s)dY_s \right|^2 &&\\[5pt] \nonumber
   &= E\left|\int_{\mathcal{T}}\int_{\mathcal{T}}\Phi^\sigma(nt-j,s_1)\Phi^\sigma(nt-j,s_2)dY_{s_1}dY_{s_2} \right|&&\\[5pt] \label{double.int.3.1}
   &= \int_{\mathcal{T}}\int_{\mathcal{T}}\left(\mathcal{L}^\sigma(nt-j)\right)^2\mathcal{L}^\sigma(s_1)\mathcal{L}^\sigma(s_2)E\left|dY_{s_1}dY_{s_2}\right|.
\end{align}
Since \( Y_s(\omega) \) has orthogonal increments, we have
\[
E|dY_{s_1}dY_{s_2}|=
\begin{cases}
 E|dY_{s_1}|^2, & \text{if } s_1=s_2, \\
  0, & \text{otherwise.}
\end{cases}
\]
Thus, by taking $s_1=s_2=s$,  the double integral in (\ref{double.int.3.1}) reduces to
\begin{align}\nonumber
     E|\varphi(nt-j,\omega)|^2 &= \int_{\mathcal{T}}\left(\mathcal{L}^\sigma(nt-j)\right)^2\left(\mathcal{L}^\sigma(s)\right)^2E\left|dY_{s}\right|^2.
\end{align}
If we denote \( \lambda_Y^2 \) as the variance of the increments of \( Y_s(\cdot) \), then we can write
\begin{align}\nonumber
     E|\varphi(nt-j,\omega)|^2 &= \lambda_Y^2 \int_{\mathcal{T}}\left(\mathcal{L}^\sigma(nt-j)\right)^2\left(\mathcal{L}^\sigma(s)\right)^2 ds &&\\[5pt] \nonumber
     &=\lambda_Y^2 \left(\mathcal{L}^\sigma(nt-j)\right)^2\left\|\mathcal{L}^\sigma(\cdot)\right\|_2.
\end{align}   

\item[(iii)] To evaluate 
\[
E\left|\int_{\mathcal{T}}\left(\frac{\Phi^\sigma(nt-j, ns-j)}{\sum\limits_{j=\lceil nc \rceil}^{\lfloor nd\rfloor-1}\Phi^\sigma(nt-j, ns-j)}\right)dY_s \right|^2,
\]
let 
\[
h(t,s) = \frac{\Phi^\sigma(nt-j, ns-j)}{\sum\limits_{j=\lceil nc \rceil}^{\lfloor nd\rfloor-1}\Phi^\sigma(nt-j, ns-j)}
\]
and 
\[
\phi_{j,n}(t,\omega) = \int_{\mathcal{T}} h(t,s) dY_s.
\]
Therefore, we have
\begin{equation}\label{double.int.3.2}
E|\phi_{j,n}(t,\omega)|^2 = \int_{\mathcal{T}}\int_{\mathcal{T}} h(t,s_1) h(t,s_2) E|dY_{s_1} dY_{s_2}|.
\end{equation}
Since \( Y_s(\omega) \) has orthogonal increments, we can express \( E|dY_{s_1} dY_{s_2}| \) as follows
\[
E|dY_{s_1} dY_{s_2}| =
\begin{cases}
E|dY_{s_1}|^2, & \text{if } s_1 = s_2, \\
0, & \text{otherwise}.
\end{cases}
\]
Thus,  by taking $s_1=s_2=s$, the double integral in  (\ref{double.int.3.2}) reduces to
\begin{align}\nonumber
    E|\phi_{j,n}(t,\omega)|^2 &= \int_{\mathcal{T}} |h(t,s)|^2 E\left|dY_{s}\right|^2&&\\[5pt] \nonumber
    & = \int_{\mathcal{T}} \frac{\left(\Phi^\sigma(nt-j, ns-j)\right)^2}{\left(\sum\limits_{j=\lceil nc \rceil}^{\lfloor nd\rfloor-1}\Phi^\sigma(nt-j, ns-j)\right)^2} E\left|dY_{s}\right|^2.&&
\end{align}

Let \(\lambda^2_Y \) represent the variance of the increments of \( Y_s(\cdot) \). Then, we obtain
\begin{align}\nonumber
E|\phi_{j,n}(t,\omega)|^2 &= \lambda^2_Y \int_{\mathcal{T}} \frac{\left(\Phi^\sigma(nt-j, ns-j)\right)^2}{\left(\sum\limits_{j=\lceil nc \rceil}^{\lfloor nd\rfloor-1}\Phi^\sigma(nt-j, ns-j)\right)^2} ds&& \\ \nonumber
& = \frac{\lambda^2_Y}{\mathcal{L}^\sigma(2)^4} \int_{\mathcal{T}} \left(\Phi^\sigma(nt-j, ns-j)\right)^2 ds&&\\[5pt] \nonumber
 & = \frac{\lambda^2_Y}{\mathcal{L}^\sigma(2)^4} \int_{\mathcal{T}} {\left(\mathcal{L}^\sigma(nt-j)\right)^2\left(\mathcal{L}^\sigma(ns-j)\right)^2ds.}
\end{align}

By changing variable $ns-j=y,$ we get
\begin{align}\nonumber
  E|\phi_{j,n}(t,\omega)|^2 & =  \frac{\lambda^2_Y}{n^2\mathcal{L}^\sigma(2)^4} \int_{\mathcal{T}} {\left(\mathcal{L}^\sigma(nt-j)\right)^2\left(\mathcal{L}^\sigma(y)\right)^2dy}&&\\[5pt] \nonumber
  & =\frac{\lambda^2_Y\left\|\mathcal{L}^\sigma(\cdot)\right\|^2_2}{n^2\mathcal{L}^\sigma(2)^4}\left(\mathcal{L}^\sigma(nt-j)\right)^2.
\end{align}
This result indicates that the expected value of \( |\phi_{j,n}(t,\omega)|^2 \) is finite.

\end{itemize}
\end{proof}

\begin{figure}[H]
    \centering
    \includegraphics[width=1\textwidth]{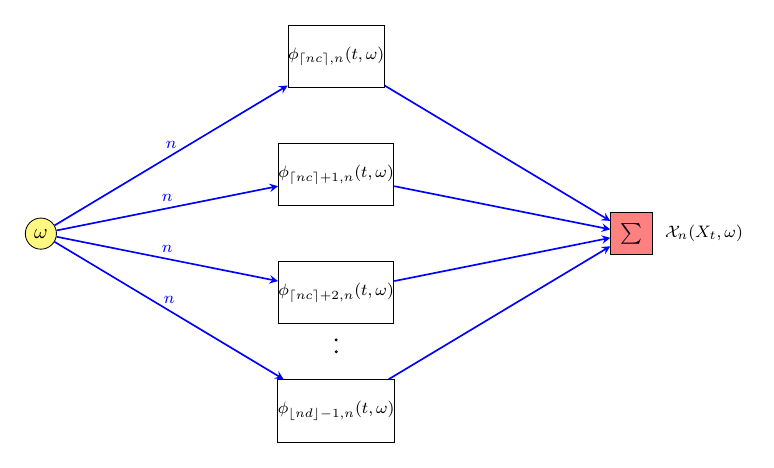} 
    \caption{Structural diagram of the K-SNNOs defined in (\ref{SNNO_by_S.neuron_definition}) }
    \label{Structural diagram of the SNNOs}
\end{figure}

\begin{definition}\label{SNNO_by_S.neuron_definition}
    Consider a stochastic process \( X_t(\omega)\in L^2(\Omega, \mathcal{F}, \mathbb{P}) \). Then, K-SNNOs  \( \mathcal{X}_n(X_t, \omega) \) activated by stochastic neurons \( \phi_{j,n}(t, \omega) \) and acting on the function $\zeta(t,s)$ related to process \( X_t(\cdot) \), are defined as follows
\begin{equation}
 \mathcal{X}_n(X_t, \omega):= \sum\limits_{j=\lceil nc \rceil}^{\lfloor nd \rfloor - 1} \left( n^2 \int_{\mathfrak{R}_{\mathbf{j},2}} \zeta(\mathbf{v}) \, d\mathbf{v} \right) \phi_{j,n}(t, \omega).
\end{equation}

Here, $\mathfrak{R}_{\mathbf{j},2}= \left[\frac{j}{n},\frac{j+1}{n}\right]\times\left[ \frac{j}{n},\frac{j+1}{n}\right]$ and \( \zeta(\cdot) \) is related to the covariance of the process, and \( \phi_{j,n}(t, \omega) \) represents the measurable stochastic neurons defined earlier. The structure of the K-SNNOs defined in \eqref{SNNO_by_S.neuron_definition} is illustrated in Figure \ref{Structural diagram of the SNNOs}.
\end{definition}

Note that, for any $X_t(\omega)\in L^2(\Omega, \mathcal{F}, \mathbb{P}),$ we get 
\begin{align}\nonumber
    E|\mathcal{X}_n(X_t,\omega)|^2 &= E\left|\sum\limits_{j=\lceil nc \rceil}^{\lfloor nd\rfloor-1}{\left(n^2\int_{\mathfrak{R}_{\mathbf{j},2}}\zeta(\mathbf{v})d\mathbf{v}\right)}{\phi_{j,n}(t,\omega)}\right|^2&&\\[5pt] \nonumber
    & \leq E\left[\sum\limits_{j=\lceil nc \rceil}^{\lfloor nd\rfloor-1}\left|{\left(n^2\int_{\mathfrak{R}_{\mathbf{j},2}}\zeta(\mathbf{v})d\mathbf{v}\right)}{\phi_{j,n}(t,\omega)}\right|^2\right]&&\\[5pt] \nonumber
    & =\sum\limits_{j=\lceil nc \rceil}^{\lfloor nd\rfloor-1}\left|n^2\int_{\mathfrak{R}_{\mathbf{j},2}}\zeta(\mathbf{v})d\mathbf{v}\right|^2E\left|{\phi_{j,n}(t,\omega)} \right|^2&&\\[5pt] \nonumber
    &  \leq \sum\limits_{j=\lceil nc \rceil}^{\lfloor nd\rfloor-1}\left(n^2\int_{\mathfrak{R}_{\mathbf{j},2}}|\zeta(\mathbf{v})|^2d\mathbf{v}\right)E\left|{\phi_{j,n}(t,\omega)} \right|^2&&\\ \nonumber 
    &\leq \frac{\lambda^2_Y\left\|\mathcal{L}^\sigma(\cdot)\right\|^2_2\left\|\zeta(\cdot)\right\|^2_2}{\mathcal{L}^\sigma(2)^4}\left(\mathcal{L}^\sigma(nt-j)\right)^2
\end{align}
in view of property $(iii)$ of Lemma \ref{Properties of S.neuron}.

Thus, for every \( t \in \mathcal{T} \), the operator \( \mathcal{X}_n(X_t, \omega) \) is well-defined and bounded in the mean square sense within the set \( L^2(\Omega, \mathcal{F}, \mathbb{P}) \).

We will now examine the convergence properties of the sequence of operators $\mathcal{X}_n(X_t,\omega)$ in the context of second-order processes and prove the following results.
\begin{theorem}
Consider a stochastic process \( X_t(\omega) \in  L^2(\Omega, \mathcal{F}, \mathbb{P}) \), indexed by \( t \in \mathcal{T} \) and admitting the canonical representation. Then,  
 $$ E\left| \mathcal{X}_n(X_t, \omega) -X_t(\omega)\right|^2\xrightarrow{} 0, \hspace{.25cm} \mbox{ for any } t\in \mathcal{T}.$$
\end{theorem}
\begin{proof}
    We have process $ X_t(\omega)\in L^2(\Omega, \mathcal{F}, \mathbb{P}),$  admitting the canonical representation 
    \begin{equation}\label{2dimKn-chi}
        X_t(\omega)= \int_{\mathcal{T}}{\zeta(t,s)dY_s},
    \end{equation}
    where $\zeta(t,s)$ is $L^2$-summable, i.e.
    $$ \int_{\mathcal{T}}\int_{\mathcal{T}}\left|\zeta(t,s)\right|^2dtds <+\infty.$$
    Now, in view of Theorem \ref{Kn-L^presult} for $2$-dimensional case, for any $\epsilon>0$, $\exists\ \mathbf{K}_n(\cdot)$ for which
    \begin{equation}
      \left\|\mathbf{K}_n(\zeta,t,s)-\zeta\right\|_2<\epsilon, \hspace{.25cm} \mbox{ for sufficiently large }n  
    \end{equation}
  and $\mathcal{X}_n(X_t, \omega):=\int_{\mathcal{T}}{\mathbf{K}_n(\zeta,t,s)}dY
  _s,$  
  we can write
    \begin{align}\nonumber
        \mathcal{X}_n(X_t, \omega)&=\int_{\mathcal{T}}{\frac{\sum\limits_{j=\lceil nc \rceil}^{\lfloor nd\rfloor-1}\mathcal{L}^\sigma(nt-j)\mathcal{L}^\sigma(ns-j)\left(n^2\int_{\mathfrak{R}_{\mathbf{j},2}}\zeta(\mathbf{v})d\mathbf{v}\right)}{\sum\limits_{j=\lceil nc \rceil}^{\lfloor nd\rfloor-1}\mathcal{L}^\sigma(nt-j)\mathcal{L}^\sigma(ns-j)}} dY_s&&\\[5pt] \nonumber
        & = \int_{\mathcal{T}}\sum\limits_{j=\lceil nc \rceil}^{\lfloor nd\rfloor-1}\left(\frac{\mathcal{L}^\sigma(nt-j)\mathcal{L}^\sigma(ns-j)}{\sum\limits_{j=\lceil nc \rceil}^{\lfloor nd\rfloor-1}\mathcal{L}^\sigma(nt-j)\mathcal{L}^\sigma(ns-j)}\right)\left(n^2\int_{\mathfrak{R}_{\mathbf{j},2}}\zeta(\mathbf{v})d\mathbf{v}\right)dY_s&&\\[5pt] \nonumber
        & =\sum\limits_{j=\lceil nc \rceil}^{\lfloor nd\rfloor-1}\left(n^2\int_{\mathfrak{R}_{\mathbf{j},2}}\zeta(\mathbf{v})d\mathbf{v}\right)\int_{\mathcal{T}}\left(\frac{\mathcal{L}^\sigma(nt-j)\mathcal{L}^\sigma(ns-j)}{\sum\limits_{j=\lceil nc \rceil}^{\lfloor nd\rfloor-1}\mathcal{L}^\sigma(nt-j)\mathcal{L}^\sigma(ns-j)}\right)dY_s&&\\ \nonumber
        & = \sum\limits_{j=\lceil nc \rceil}^{\lfloor nd\rfloor-1}\left(n^2\int_{\mathfrak{R}_{\mathbf{j},2}}\zeta(\mathbf{v})d\mathbf{v}\right)\phi_{j,n}(t,\omega),
        \end{align}
        where $\int_{\mathcal{T}}\left(\frac{\mathcal{L}^\sigma(nt-j)\mathcal{L}^\sigma(ns-j)}{\sum\limits_{j=\lceil nc \rceil}^{\lfloor nd\rfloor-1}\mathcal{L}^\sigma(nt-j)\mathcal{L}^\sigma(ns-j)}\right)dY_s= \phi_{j,n}(t,\omega)$.
       
     Since canonical representation establishes an isometry between $L^2(\mathcal{T}\times \mathcal{T})$ and $L^2(\Omega, \mathcal{F}, \mathbb{P})$, the following equality holds
    \begin{equation}\label{iso.1}
        E\left\{\left|\mathcal{X}_n(X_t, \omega) -X_t(\omega)\right|^2\right\}= \int_{\mathcal{T}}\left|\mathbf{K}_n(\zeta,t,s)-\zeta(t,s)\right|^2ds
    \end{equation}
     from (\ref{2dimKn-chi}) and (\ref{iso.1}), we conclude that the stochastic process
     $$\mathcal{X}_n(X_t, \omega)= \int_{\mathcal{T}}{\mathbf{K}_n(\zeta,t,s)}dY_s $$
    can approximate given process $X_t(\omega)$ in mean square sense.
\end{proof}

We will now examine the quantitative estimates for the sequence of operators $\mathcal{X}_n(X_t,\omega)$ in the space $L^2 (\Omega, \mathcal{F}, \mathbb{P})$ and measure the rate of approximation error in terms of modulus of continuity defined below.

Let $ \zeta(t,s)\in L^2 (\mathcal{T}\times\mathcal{T})$. For $\delta>0$, the function 
 \[
\mathcal{W}_2(\zeta, \delta) = \sup_{|v_1 - t|,|v_2 - s| \leq \delta} \left( \int_{\mathcal{T}} \int_{\mathcal{T}} |\zeta(v_1, v_2) - \zeta(t,s)|^2 \,dv_1 \,dv_2 \right)^{\frac{1}{2}}
\]
 is known as the $ L^2$-modulus of continuity of $\zeta(t,s).$ We write $Lip_2(u)=\{ \zeta\in L^2 (\mathcal{T}\times\mathcal{T}): \mathcal{W}_2(\zeta, \delta)=\mathcal{O}(\delta^u); 0<u\leq 1\}$.

\begin{theorem}\label{Theorem-L^2-quantitative}
 Let \( X_t(\omega) \in L^2 (\Omega, \mathcal{F}, \mathbb{P})\),  \(\zeta(t,s)\in  Lip_2(u)\) and let \( \mathcal{X}_n(X_t, \omega) \) be its approximation. Assume that the sigmoidal density function $\mathcal{L}^\sigma(\cdot)$ satisfies, for some $\alpha,\theta>0$, \(\int_{x\leq \frac{1}{n^\alpha}}n^2\mathcal{L}^\sigma(nx)x^{2u}dx=\mathcal{O}(n^{-\theta})\) as \(n\to+\infty.\) Then, the mean-square error (MSE) satisfies 
  \[
E\left| X_t(\omega) - \mathcal{X}_n(X_t, \omega) \right|^2 =\mathcal{O}(n^{-\mu}),\]
 where \(\mu= \min\{\theta,\beta>0,2u\}.\)  
\end{theorem}

\begin{proof} Since $X_t(\omega)\in L^2(\Omega,\mathcal F,\mathbb P)$ admits the canonical representation. Using conditions $(iii) \mbox{ and } (vi) $ of Lemma \ref{lemma-density_fun. properties}, the MSE satisfies, 
by the stochastic-integral isometry,
{\footnotesize
\begin{align}\nonumber
   &E\left|e_n(t,\omega)\right|^2=E\left|\mathcal{X}_n(X_t, \omega)-X_t(\omega)  \right|^2&&\\[5pt] \nonumber
   &\qquad\qquad\quad = \int_{\mathcal{T}}\left|\mathbf{K}_n(\zeta,t,s)-\zeta(t,s)\right|^2ds&&\\[5pt] \nonumber
   & = \int_{\mathcal{T}}\left|{\frac{\sum\limits_{j=\lceil nc \rceil}^{\lfloor nd\rfloor-1}\mathcal{L}^\sigma(nt-j)\mathcal{L}^\sigma(ns-j)\left(n^2\int_{\mathfrak{R}_{\mathbf{j},2}}\zeta(\mathbf{v})d\mathbf{v}\right)}{\sum\limits_{j=\lceil nc \rceil}^{\lfloor nd\rfloor-1}\mathcal{L}^\sigma(nt-j)\mathcal{L}^\sigma(ns-j)}} -\zeta(t,s)\right|^2ds&&\\[5pt] \nonumber
   &\leq  \frac{1}{\mathcal{L}^\sigma(2)^4}\int_{\mathcal{T}}\left|{\sum\limits_{j=\lceil nc \rceil}^{\lfloor nd\rfloor-1}\mathcal{L}^\sigma(nt-j)\mathcal{L}^\sigma(ns-j)\left(n^2\int_{\mathfrak{R}_{\mathbf{j},2}}[\zeta(\mathbf{v})-\zeta(t,s)]d\mathbf{v}\right)}\right|^2ds&&\\[5pt] \nonumber
  &\leq  \frac{\mathcal{M}_0(\mathcal{L}^\sigma)}{\mathcal{L}^\sigma(2)^4}\int_{\mathcal{T}}{\sum\limits_{j=\lceil nc \rceil}^{\lfloor nd\rfloor-1}\mathcal{L}^\sigma(ns-j)\left(n^2\int_{\mathfrak{R}_{\mathbf{j},2}}\left|\zeta(\mathbf{v}) -\zeta(t,s)\right|^2d\mathbf{v}\right)}ds&&\\[5pt] \nonumber &\leq  \frac{\mathcal{M}_0(\mathcal{L}^\sigma)}{\mathcal{L}^\sigma(2)^4}\int_{\mathcal{T}}{\sum\limits_{j=\lceil nc \rceil}^{\lfloor nd\rfloor-1}\mathcal{L}^\sigma(ns-j)\left(n^2\int_{\mathfrak{R}_{\mathbf{j},2}}\left|\zeta(v_1,v_2) -\zeta(t,s)\right|^2dv_1dv_2\right)}ds&&\\[5pt] \nonumber 
  &\leq \frac{\mathcal{M}_0(\mathcal{L}^\sigma)}{\mathcal{L}^\sigma(2)^4}\int_{\mathcal{T}}\sum\limits_{j=\lceil nc \rceil}^{\lfloor nd\rfloor-1}\mathcal{L}^\sigma(ns-j)\Bigl(n^2\int_{\mathfrak{R}_{\mathbf{j},2}}\Bigl|\zeta(v_1,v_2) - \zeta(v_1+s-\frac{j}{n},v_2+s-\frac{j}{n})&&\\[5pt] \nonumber
 & \qquad\qquad\qquad\qquad\qquad\qquad+\zeta(v_1+s-\frac{j}{n},v_2+s-\frac{j}{n})-\zeta(t,s)\Bigl|^2dv_1dv_2\Bigl)ds&&\\[5pt] \nonumber 
 &\leq \frac{\mathcal{M}_0(\mathcal{L}^\sigma)}{\mathcal{L}^\sigma(2)^4}\int_{\mathcal{T}}\sum\limits_{j=\lceil nc \rceil}^{\lfloor nd\rfloor-1}\mathcal{L}^\sigma(ns-j)\Bigl(n^2\int_{\mathfrak{R}_{\mathbf{j},2}}\Bigl|\zeta(v_1,v_2)&&\\[5pt] \nonumber
 &\qquad\qquad\qquad\qquad\qquad\qquad\qquad\qquad\qquad -\zeta(v_1+s-\frac{j}{n},v_2+s-\frac{j}{n})\Bigl|^2 dv_1dv_2&&\\[5pt] \nonumber
 & \qquad\qquad\qquad\qquad+n^2\int_{\mathfrak{R}_{\mathbf{j},2}}\Bigl|\zeta(v_1+s-\frac{j}{n},v_2+s-\frac{j}{n})-\zeta(t,s)\Bigl|^2dv_1dv_2\Bigl)ds&& \\ \nonumber 
 & = I_1+I_2.
\end{align}}
Let us first estimate \(I_1\). By applying Jensen's inequality and changing the variable  \(x = s - \frac{j}{n}\), and for some \(\alpha>0\), we obtain
{\footnotesize
\begin{align}\nonumber
    I_1= & \frac{\mathcal{M}_0(\mathcal{L}^\sigma)}{\mathcal{L}^\sigma(2)^4}\int_{\mathcal{T}}\sum\limits_{j=\lceil nc \rceil}^{\lfloor nd\rfloor-1}\mathcal{L}^\sigma(ns-j)\Bigl(n^2\int_{\mathfrak{R}_{\mathbf{j},2}}\Bigl|\zeta(v_1,v_2)&&\\[5pt] \nonumber
    &\qquad\qquad\qquad\qquad\qquad\qquad\qquad- \zeta(v_1+s-\frac{j}{n},v_2+s-\frac{j}{n})\Bigl|^2 dv_1dv_2\Bigl)ds&&\\[5pt] \nonumber
    =& \frac{\mathcal{M}_0(\mathcal{L}^\sigma)}{\mathcal{L}^\sigma(2)^4}\int_{\mathcal{T}}n^2\mathcal{L}^\sigma(nx)\Bigl(\sum\limits_{j=\lceil nc \rceil}^{\lfloor nd\rfloor-1}\int_{\mathfrak{R}_{\mathbf{j},2}}\Bigl|\zeta(v_1,v_2) - \zeta(v_1+x,v_2+x)\Bigl|^2 dv_1dv_2\Bigl)dx&&\\[5pt] \nonumber
     \leq&\frac{\mathcal{M}_0(\mathcal{L}^\sigma)}{\mathcal{L}^\sigma(2)^4}\int_{\mathcal{T}}n^2\mathcal{L}^\sigma(nx)\Bigl(\int_{\mathcal{T}}\int_{\mathcal{T}}\Bigl|\zeta(v_1,v_2) - \zeta(v_1+x,v_2+x)\Bigl|^2 dv_1dv_2\Bigl)dx&& \\[5pt] \nonumber
    \leq&\frac{\mathcal{M}_0(\mathcal{L}^\sigma)n^2}{\mathcal{L}^\sigma(2)^4}\left[\int_{0\leq x\leq \frac{1}{n^\alpha}}+\int_{x>\frac{1}{n^\alpha}}\right]\mathcal{L}^\sigma(nx) \Bigl(\int_{\mathcal{T}}\int_{\mathcal{T}}\Bigl|\zeta(v_1,v_2) - \zeta(v_1+x,v_2+x)\Bigl|^2 dv_1dv_2\Bigl)dx&&\\[5pt] \nonumber
   =&J_1 +J_2.
\end{align}}
For $J_1$, since \(\zeta(t,s)\in Lip_2(u)\), we have
\begin{align}\nonumber
    J_1 \leq& \frac{\mathcal{M}_0(\mathcal{L}^\sigma)}{\mathcal{L}^\sigma(2)^4}\int_{0\leq x\leq \frac{1}{n^\alpha}}n^2\mathcal{L}^\sigma(nx)\Bigl(\int_{\mathcal{T}}\int_{\mathcal{T}}\Bigl|\zeta(v_1,v_2) - \zeta(v_1+x,v_2+x)\Bigl|^2 dv_1dv_2\Bigl)dx&&\\[5pt] \label{J_1}
    \leq& \frac{\mathcal{M}_0(\mathcal{L}^\sigma)}{\mathcal{L}^\sigma(2)^4}\int_{0\leq x\leq \frac{1}{n^\alpha}}n^2\mathcal{L}^\sigma(nx)x^{2u}dx=\mathcal{O}(n^{-\theta}) \mbox{ as } n\to+\infty.
\end{align}
For $J_2,$ we have
\begin{align}\nonumber
    J_2 \leq& \frac{\mathcal{M}_0(\mathcal{L}^\sigma)n^2}{\mathcal{L}^\sigma(2)^4}\int_{x>\frac{1}{n^\alpha}}\mathcal{L}^\sigma(nx)\Bigl(\int_{\mathcal{T}}\int_{\mathcal{T}}\Bigl|\zeta(v_1,v_2) - \zeta(v_1+x,v_2+x)\Bigl|^2 dv_1dv_2\Bigl) dx &&\\[5pt] \nonumber
     \leq& \frac{\mathcal{M}_0(\mathcal{L}^\sigma)n^2}{\mathcal{L}^\sigma(2)^4}\int_{x>\frac{1}{n^\alpha}}\mathcal{L}^\sigma(nx)\left\{\Bigl\|\zeta(\cdot,\cdot)\Bigl\|^2 + \Bigl\|\zeta(\cdot+x,\cdot+x)\Bigl\|_2^2\right\} dx &&\\[5pt] \nonumber 
      \leq& \frac{2\mathcal{M}_0(\mathcal{L}^\sigma)\Bigl\|\zeta(\cdot,\cdot)\Bigl\|_2^2}{\mathcal{L}^\sigma(2)^4}\int_{x>\frac{1}{n^\alpha}}n^2\mathcal{L}^\sigma(nx)dx.
     \end{align}
By change of variables $y=nx$, we have
\[
\int_{x>1/n^\alpha} n^2\mathcal L^\sigma(nx)\,dx
= n\int_{y>n^{1-\alpha}}\mathcal L^\sigma(y)\,dy.
\]
Using (P3), there are $\gamma>0$ and $C>0$ such that 
$\mathcal L^\sigma(y)\le C y^{-1-\gamma}$ for large $y$. \\
Hence for large $n$, we have
\[
n\int_{y>n^{1-\alpha}}\mathcal L^\sigma(y)\,dy
\le Cn\int_{n^{1-\alpha}}^\infty y^{-1-\gamma}\,dy
=\frac{C}{\gamma}\,n^{\,1-\gamma(1-\alpha)}.
\]
Set $\beta:=\gamma(1-\alpha)-1$. Then
\begin{equation}\label{eq:J_2}
\int_{x>1/n^\alpha} n^2\mathcal L^\sigma(nx)\,dx
= \mathcal O(n^{-\beta}).    
\end{equation}

Now, we can estimate $I_2$. Using Jensen's inequality, the change of variables $y_1=v_1-\frac{j}{n},\; y_2=v_2-\frac{j}{n}$, and Fubini-Tonelli theorem, we have
{\footnotesize
\begin{align}\nonumber
    I_2= &\frac{\mathcal{M}_0(\mathcal{L}^\sigma)}{\mathcal{L}^\sigma(2)^4}\int_{\mathcal{T}}\sum\limits_{j=\lceil nc \rceil}^{\lfloor nd\rfloor-1}\mathcal{L}^\sigma(ns-j)\Bigl(n^2\int_{\mathfrak{R}_{\mathbf{j},2}}\Bigl|\zeta(v_1+s-\frac{j}{n},v_2+s-\frac{j}{n})&&\\[5pt] \nonumber
&\qquad\qquad\qquad\qquad\qquad\qquad\qquad\qquad\qquad\qquad\qquad\qquad-\zeta(t,s)\Bigl|^2dv_1dv_2\Bigl)ds&&\\[5pt] \nonumber
   \leq & \frac{\mathcal{M}^2_0(\mathcal{L}^\sigma)}{\mathcal{L}^\sigma(2)^4}\int_{\mathcal{T}}\Bigl(n^2\int_{0}^{\frac{1}{n}}\int_{0}^{\frac{1}{n}}\Bigl|\zeta(s+y_1, s+y_2)-\zeta(t,s)\Bigl|^2dy_1dy_2\Bigl)ds&&\\[5pt] \nonumber
   \leq & \frac{\mathcal{M}^2_0(\mathcal{L}^\sigma)}{\mathcal{L}^\sigma(2)^4}\Bigl(n^2\int_{0}^{\frac{1}{n}}\int_{0}^{\frac{1}{n}}\Bigl\|\zeta(\cdot+y_1, \cdot+y_2)-\zeta(\cdot,\cdot)\Bigl\|_2^2dy_1dy_2\Bigl)&&\\[5pt] \nonumber
   \leq & \frac{\mathcal{M}^2_0(\mathcal{L}^\sigma)\mathcal{W}^2_2(\zeta,\frac{1}{n})}{\mathcal{L}^\sigma(2)^4}\Bigl(n^2\int_{0}^{\frac{1}{n}}\int_{0}^{\frac{1}{n}}dy_1dy_2\Bigl)&&\\[5pt] \nonumber
 \leq & \frac{\mathcal{M}^2_0(\mathcal{L}^\sigma)\mathcal{W}^2_2(\zeta,\frac{1}{n})}{\mathcal{L}^\sigma(2)^4}.  
\end{align}}
Since $\zeta\in\mathrm{Lip}_2(u)$,
$\mathcal W_2^2(\zeta,\delta)=\mathcal O(\delta^{2u})$, and thus
\begin{equation}\label{I_2}
 I_2
   \le \frac{\mathcal M_0^2(\mathcal L^\sigma)}{\mathcal{L}^\sigma(2)^4}
        \mathcal W_2^2\!\left(\zeta,\tfrac{1}{n}\right)
   =\mathcal O(n^{-2u}).   
\end{equation}

Combining all estimates (\ref{J_1}), (\ref{eq:J_2}) and (\ref{I_2}), we obtain
\[
E\!\left|X_t(\omega)-\mathcal X_n(X_t,\omega)\right|^2
   =\mathcal O(n^{-\mu}),
   \qquad
   \mu=\min\{\theta,\beta,2u\}.
\]
This completes the proof.
\end{proof}

\section{Numerical validation}
For numerical validation of our theoretical results, we first specify a second-order stochastic process $\{X_t(\omega)\}$ and then determine the corresponding deterministic kernel $\zeta(t,s)$ that generates the process through the Wiener integral representation
\[
    X_t(\omega) = \int_{\mathcal{T}} \zeta(t,s)\, dW_s(\omega).
\]
Consider the process
\[
    X_t(\omega)= t^2\Bigg[W(1)-\int_{0}^{1}W_s\,ds\Bigg], \qquad t\in[0,1],
\]
which belongs to $L^2(\Omega, \mathcal{F}, \mathbb{P})$ for every $t$.  
The associated kernel $\zeta(t,s)$ satisfying
\[
    X_t(\omega) = \int_{\mathcal{T}}\zeta(t,s)\,dW_s
\]
is obtained as
\[
    \zeta(t,s)=
    \begin{cases}
        t^2 s, & s\in [0,1],\\[4pt]
        0,     & \text{otherwise}.
    \end{cases}
\]
Hence,
\begin{align}\label{OriganX_teq.}
    X_t(\omega)
        &= \int_{0}^{1} t^2 s \, dW_s(\omega)
         = t^2\left[W(1)-\int_{0}^{1} W_s\,ds\right].
\end{align}
For details about the above stochastic integral, one can refer \cite{brzezniak2000basic-stochastic}.

This representation enables us to compute the error $E\left|\mathcal{X}_n(X_t, \omega)-X_t(\omega)  \right|^2$ theoretically as well as numerically.

\begin{figure}[H]
    \centering
    \includegraphics[width=0.80\linewidth]{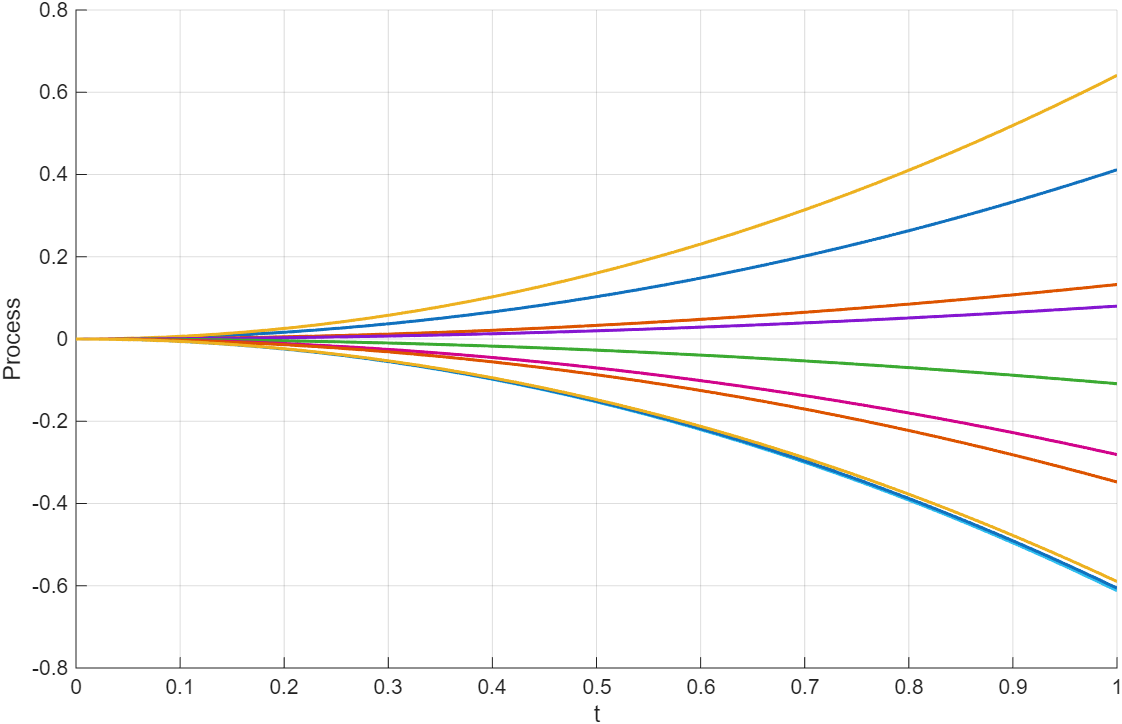}
    \caption{10 realizations of the process $X_t(\cdot)$  as defined in (\ref{OriganX_teq.})}
    \label{OrignalX_tgraph}
\end{figure}

Some realizations of process $(\ref{OriganX_teq.})$ w.r.t. a fixed realization of $dW_s$ is given in the below Figure $\ref{OrignalX_tgraph}$.

Here, $ E{|X_t(\omega)|}=0$, so that for $t,s\in[0,1],$ we have
\begin{align}\nonumber
    E|X_t(\omega)|^2&=t^4 E\left[\left(\int_{0}^{1}sdW_s\right)^2\right]&&\\[5pt] \nonumber
    & = t^4\int_{0}^{1}s^2ds= \frac{t^4}{3},\end{align} 
and \begin{align}\nonumber
  E\left|X_t(\omega)X_s(\omega)\right|&=E\left|t^2s^2\int_{0}^{1}\int_{0}^{1}{v_1v_2} dW_{v_1} dW_{v_2}\right|&&\\[5pt] \nonumber
    & = t^2s^2\int_{0}^{1}{v_1^2}dv_1=\frac{(ts)^2}{3}.
\end{align}

Now, taking $\sigma(t)=\frac{1}{1+e^{-t}};\hspace{.25cm}t\in \mathcal{T}=[0,1]$, we have
$$ \mathcal{L}^\sigma(t)=\frac{1}{2}[\sigma(t+1)-\sigma(t-1)]$$
and 
\begin{equation}\label{Snuron.2}
    \phi_{j,n}(t,\omega)= \int_{0}^{1}\left(\frac{\mathcal{L}^\sigma(nt-j)\mathcal{L}^\sigma(ns-j)}{\sum\limits_{j=0}^{n-1}\mathcal{L}^\sigma(nt-j)\mathcal{L}^\sigma(ns-j)}\right)dW_s ;\hspace{.25cm} t,s \in [0,1].
\end{equation}

Then, approximation of the above defined stochastic process (\ref{OriganX_teq.}), by K-SNNO (\ref{SNNO_by_S.neuron_definition})
is as under.
\begin{align}\nonumber
  \mathcal{X}_n(X_t, \omega) =& \sum\limits_{j=0}^{n-1} \left( n^2 \int_{\frac{j}{n}}^{\frac{j+1}{n}}\int_{\frac{j}{n}}^{\frac{j+1}{n}} f(v_1,v_2) \, {dv_1dv_2} \right) \phi_{j,n}(t, \omega)&&\\[5pt] \label{approximation4.2}
   =& \sum\limits_{j=0}^{n-1} \left( n^2 \int_{\frac{j}{n}}^{\frac{j+1}{n}}\int_{\frac{j}{n}}^{\frac{j+1}{n}} v_1^2v_2 \, {dv_1dv_2} \right) \phi_{j,n}(t, \omega),\quad t \in \mathcal{T}.
\end{align}

\begin{figure}[H]
    \centering
    \includegraphics[width=0.8\linewidth]{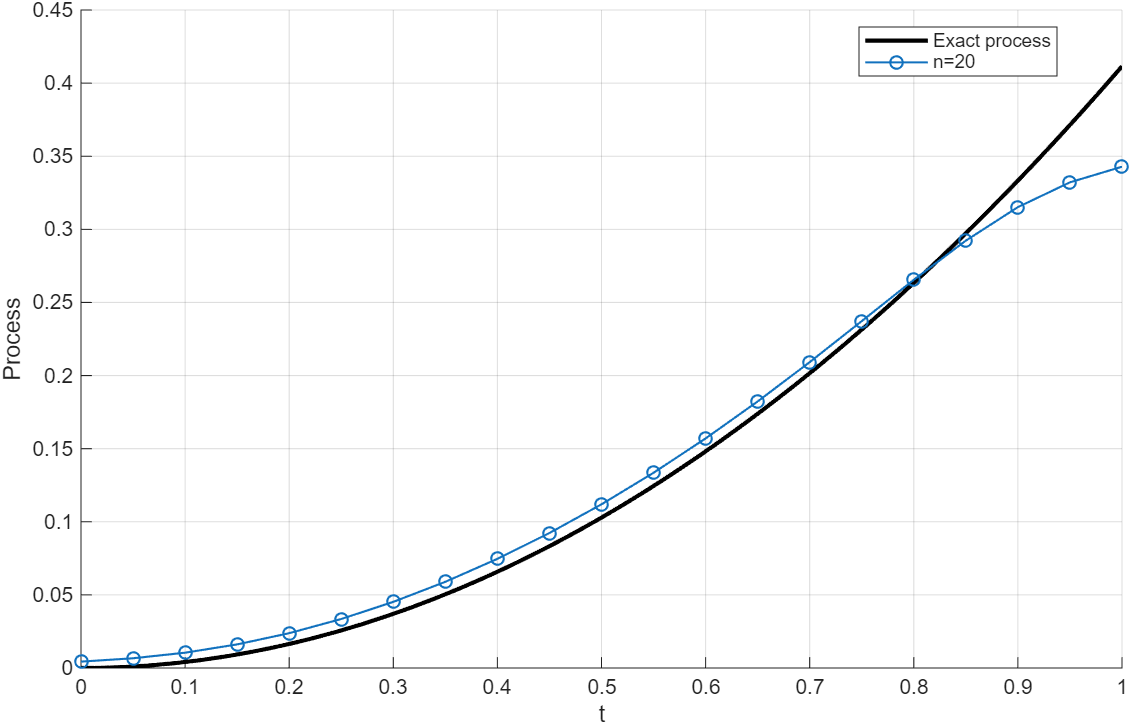}
    \caption{One realization of the approximation \((\ref{approximation4.2})\) corresponding to a fixed realization of \( dW_s \) for $n=20.$ }
    \label{fig:KSNNO(n=20)}
\end{figure}

One realization of the approximation \((\ref{approximation4.2})\) corresponding to a fixed realization of \( dW_s \) for $n=20$ are depicted in Figure \(\ref{fig:KSNNO(n=20)}\).
Furthermore, Table \ref{table-1} provides corresponding numerical data for \( n = 20 \), showcasing the comparison between actual values and approximated values.
\begin{table}[H]
\centering
\begin{tabular}{|c|c|c|c|c|c|c|}
\hline
$t_i$ & $dW_{t_i}$ & $W_{t_i}$ & $X_{t_i}$ & $\mathcal{X}_n(X_{t_i})$&  Abs. Error & Sq. Error \\
\hline
0.000 &  0.0000 &  0.0000 & 0.0000 & 0.0044 & 0.0044 & 0.000019 \\
0.050 &  0.0714 &  0.0714 & 0.0010 & 0.0066 & 0.0056 & 0.000032 \\
0.100 & -0.0902 & -0.0188 & 0.0041 & 0.0105 & 0.0063 & 0.000040 \\
0.150 &  0.0586 &  0.0399 & 0.0093 & 0.0161 & 0.0069 & 0.000047 \\
0.200 & -0.0317 &  0.0082 & 0.0165 & 0.0238 & 0.0073 & 0.000054 \\
0.250 & -0.2241 & -0.2159 & 0.0257 & 0.0335 & 0.0078 & 0.000061 \\
0.300 &  0.0306 & -0.1853 & 0.0370 & 0.0452 & 0.0082 & 0.000067 \\
0.350 & -0.1392 & -0.3244 & 0.0504 & 0.0589 & 0.0085 & 0.000073 \\
0.400 & -0.0348 & -0.3593 & 0.0658 & 0.0746 & 0.0088 & 0.000078 \\
0.450 & -0.3190 & -0.6783 & 0.0833 & 0.0923 & 0.0090 & 0.000081 \\
0.500 &  0.1985 & -0.4798 & 0.1029 & 0.1120 & 0.0091 & 0.000082 \\
0.550 & -0.1134 & -0.5932 & 0.1245 & 0.1335 & 0.0090 & 0.000082 \\
0.600 &  0.1818 & -0.4114 & 0.1481 & 0.1570 & 0.0089 & 0.000079 \\
0.650 &  0.2563 & -0.1551 & 0.1739 & 0.1823 & 0.0084 & 0.000071 \\
0.700 &  0.3746 &  0.2195 & 0.2016 & 0.2092 & 0.0076 & 0.000057 \\
0.750 & -0.3119 & -0.0924 & 0.2315 & 0.2372 & 0.0058 & 0.000033 \\
0.800 &  0.0381 & -0.0543 & 0.2634 & 0.2655 & 0.0021 & 0.000005 \\
0.850 & -0.0914 & -0.1457 & 0.2973 & 0.2923 & 0.0050 & 0.000025 \\
0.900 & -0.0063 & -0.1520 & 0.3333 & 0.3152 & 0.0181 & 0.000328 \\
0.950 & -0.0124 & -0.1645 & 0.3714 & 0.3321 & 0.0392 & 0.001538 \\
1.000 &  0.3964 &  0.2320 & 0.4115 & 0.3428 & 0.0687 & 0.004716 \\
\hline
\end{tabular}
\caption{Corresponding numerical data for $n=20.$}
\label{table-1}
\end{table}

\begin{figure}[H]
    \centering
    \includegraphics[width=0.8\linewidth]{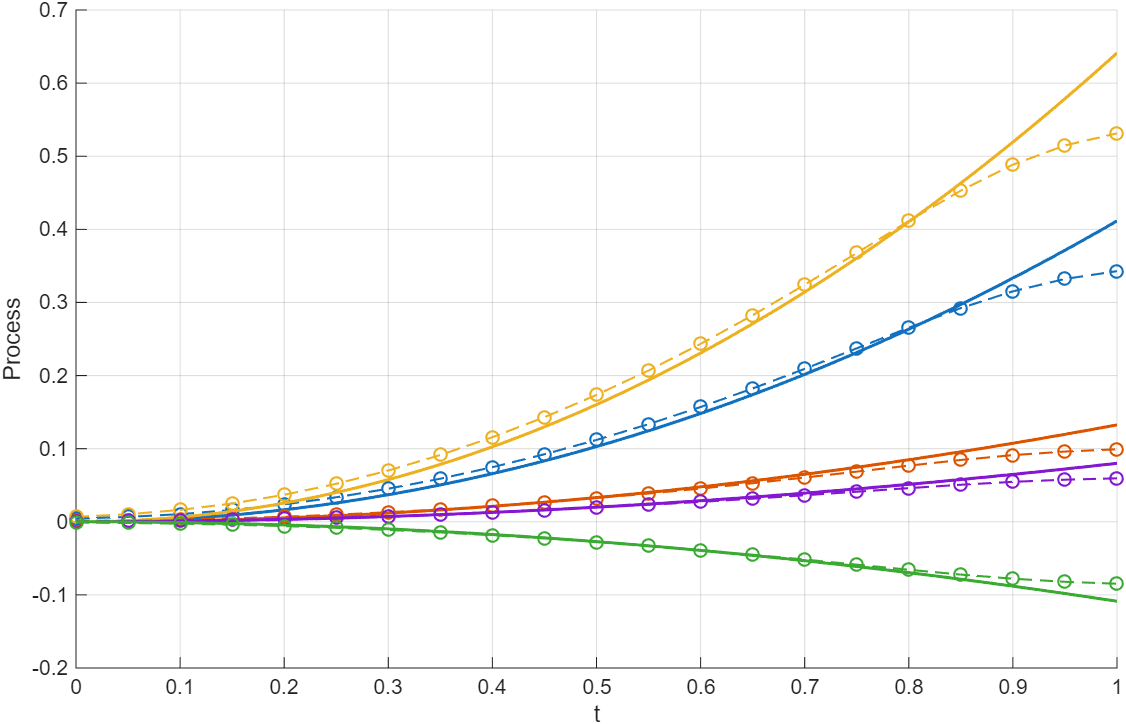}
    \caption{Comparison between the 5 realizations of original stochastic process $X_t(\omega)$ (\ref{OriganX_teq.})
 (solid lines) and its corresponding approximations via K-SNNO \eqref{approximation4.2} (dashed lines) for fixed  $n=20$.}
    \label{fig:KSNNO_multirealizations(n=20)}
\end{figure}

\begin{figure}[H]
    \centering
    \includegraphics[width=0.8\linewidth]{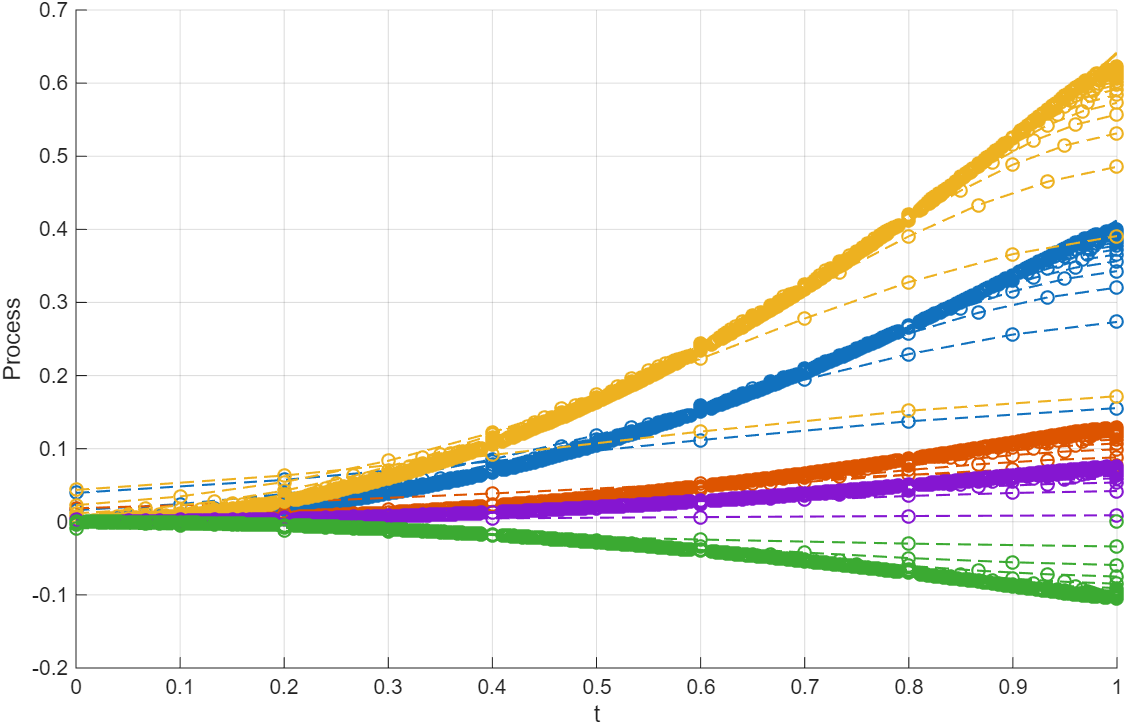}
    \caption{Comparison between the 5 realizations of original stochastic process $X_t(\omega)$ (\ref{OriganX_teq.})
 (solid lines) and its corresponding approximations via K-SNNO \eqref{approximation4.2} (dashed lines) for  $n=5,10,15,....,100$.}
    \label{fig:KSNNO_5_realization(n=0-5-100)}
\end{figure}

Several realizations of the approximation (\ref{approximation4.2}) corresponding to a realization of \( dW_s \) are depicted in Figures \ref{fig:KSNNO_multirealizations(n=20)} and \ref{fig:KSNNO_5_realization(n=0-5-100)}.

In Figures \ref{fig:KSNNO_multirealizations(n=20)} and \ref{fig:KSNNO_5_realization(n=0-5-100)} the approximations closely follow the true trajectories, demonstrating that K-SNNO effectively preserves the stochastic behaviour of the process. Across all realizations, the approximation captures the general trend and curvature without introducing oscillations. Small deviations are observed around the mid-interval, likely due to the localization behaviour of the activation function $ \phi_{j,n}(t,\omega)$. Overall, the results confirm the smoothness, consistency, and accuracy of K-SNNO-based approximations.

\begin{figure}[H]
    \centering
    \includegraphics[width=0.8\linewidth]{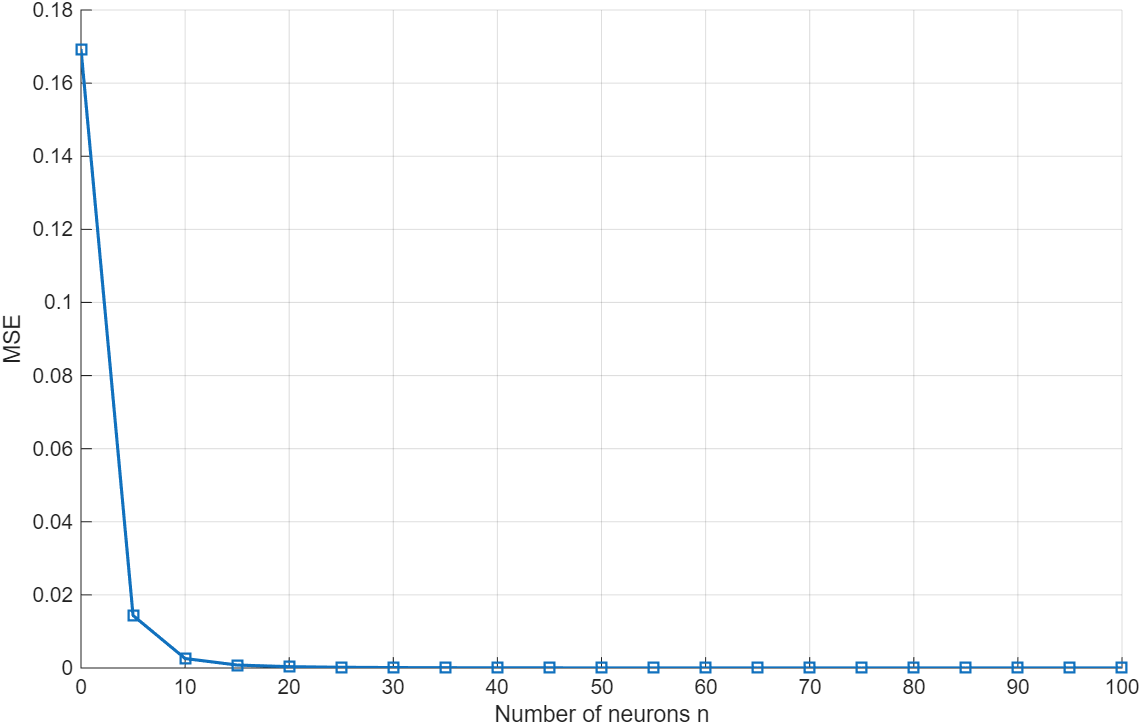}
    \caption{Show how MSE decreases w.r.t. $n$ for fixed realization.}
    \label{MSE-plot-1}
\end{figure}
\begin{figure}[H]
    \centering
    \includegraphics[width=0.8\linewidth]{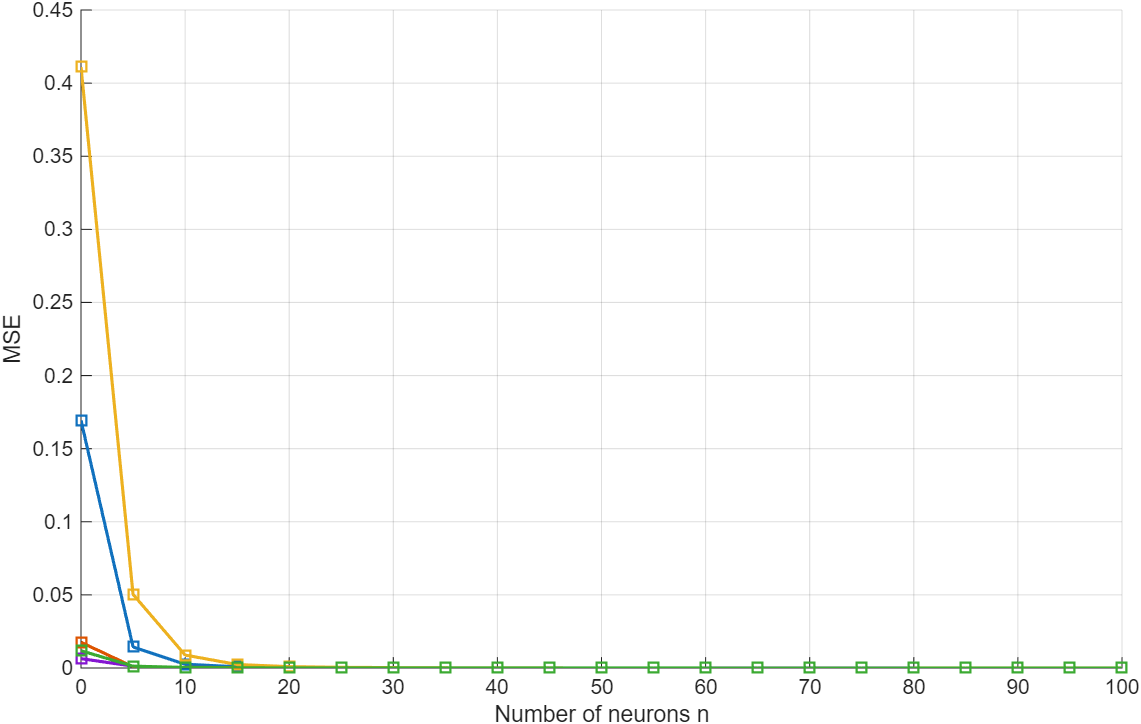}
    \caption{Show how MSE decreases w.r.t. $n$ for different  realization.}
    \label{MSE-plot-2}
\end{figure}

The plotted curve demonstrates that the Mean Squared Error (MSE) of the K-SNNO approximation decreases consistently as the discretization parameter \( n \) increases. This confirms the theoretical convergence behaviour predicted by the bound
\[
{E}\left| X_t(\omega) - \mathcal{X}_n(X_t, \omega) \right|^2 \leq \mathcal{O}(n^{-\mu}),
\]
with \( \mu > 0 \) depending on the smoothness and localization properties of the activation and the process covariance structure. The logarithmic scale emphasizes that beyond \( n \approx 100 \), the MSE stabilizes, suggesting that higher resolution provides diminishing returns due to the bounded stochastic fluctuation and kernel saturation effects. This validates the effectiveness of the K-SNNO operator in approximating second-order stochastic processes with controlled error rates.

\section{Conclusion}

In this work, we introduce Kantorovich-type stochastic neural network operators (K-SNNOs) for mean square approximation of stochastic processes. Using suitable activation functions, we established their theoretical foundations and validated their performance through numerical simulations. Graphical results demonstrate that the approximation error decreases with more neurons, confirming the accuracy and efficiency of K-SNNOs. Future directions include extending the framework to multidimensional processes, exploring various activation functions, and analyzing the trade-off between computational cost and accuracy for practical applications.

\textbf{\large Declaration of Competing Interest}

The authors affirm that there are no financial or personal affiliations that could have influenced the content or outcomes presented in this manuscript.

\bibliographystyle{unsrt}  
\bibliography{Ref}

@article{funahashi1989approximate,
  title={On the approximate realization of continuous mappings by neural networks},
  author={Funahashi, Ken-Ichi},
  journal={Neural Netw.},
  volume={2},
  number={3},
  pages={183--192},
  year={1989},
  publisher={Elsevier}
}

@article{cybenko1989approximation,
  title={Approximation by superpositions of a sigmoidal function},
  author={Cybenko, George},
  journal= {Math. Control Signals Syst.},
  volume={2},
  number={4},
  pages={303--314},
  year={1989},
  publisher={Springer}
}

@article{hornik1989multilayer,
  title={Multilayer feedforward networks are universal approximators},
  author={Hornik, Kurt and Stinchcombe, Maxwell and White, Halbert},
  journal={Neural Netw.},
  volume={2},
  number={5},
  pages={359--366},
  year={1989},
  publisher={Elsevier}
}

@misc{doob1990stochastic,
  title={Stochastic Processes. Wiley Classics Library},
  author={Doob, JL},
  year={1990},
  publisher={Wiley New York}
}

@article{amari1992information-boltezman,
  title={Information geometry of Boltzmann machines},
  author={Amari, Shun ichi and Kurata, Koji and Nagaoka, Hiroshi},
  journal={IEEE Trans.  Neural Netw.},
  volume={3},
  number={2},
  pages={260--271},
  year={1992},
  publisher={IEEE}
}

@article{conti1994approximation,
  title={Approximation of dynamical systems by continuous-time recurrent approximate identity neural networks},
  author={Conti, Massimo and Turchetti, Claudio},
  journal={Neural, Parallel \& Sci. Comput.},
  volume={2},
  number={3},
  pages={299--320},
  year={1994},
  publisher={Dynamic Publishers, Inc. Atlanta, GA, USA}
}

@inproceedings{zhao1996recurrent-SNN,
  title={A recurrent network with stochastic weights},
  author={Zhao, Jieyu and Shawe-Taylor, John},
  booktitle={Proceedings of International Conference on Neural Networks (ICNN'96)},
  volume={2},
  pages={1302--1307},
  year={1996},
  organization={IEEE}
}

@article{makovoz1996random-APP.with-NN,
  title={Random approximants and neural networks},
  author={Makovoz, Yuly},
  journal={J. Approx. Theory},
  volume={85},
  number={1},
  pages={98--109},
  year={1996},
  publisher={Elsevier}
}

@article{belli1999artificialNN,
  title={Artificial neural networks as approximators of stochastic processes},
  author={Belli, Marcello R and Conti, Massimo and Crippa, Paolo and Turchetti, Claudio},
  journal={Neural Netw.},
  volume={12},
  number={4-5},
  pages={647--658},
  year={1999},
  publisher={Elsevier}
}

@article{cardaliaguet1992approximation,
  title={Approximation of a function and its derivative with a neural network},
  author={Cardaliaguet, Pierre and Euvrard, Guillaume},
  journal={Neural Netw.},
  volume={5},
  number={2},
  pages={207--220},
  year={1992},
  publisher={Elsevier}
}

@article{anastassiou1997rate,
  title={Rate of convergence of some neural network operators to the unit-univariate case},
  author={Anastassiou, George A},
  journal={J. Math. Anal. Appl.},
  volume={212},
  number={1},
  pages={237--262},
  year={1997},
  publisher={Elsevier}
}

@book{brzezniak2000basic-stochastic,
  title={Basic stochastic processes: a course through exercises},
  author={Brzezniak, Zdzislaw and Zastawniak, Tomasz},
  year={2000},
  publisher={Springer Science \& Business Media}
}

@article{anastassiou2000rate,
  title={Rate of convergence of some multivariate neural network operators to the unit},
  author={Anastassiou, George A},
  journal={Comput.  Math. Appl.},
  volume={40},
  number={1},
  pages={1--19},
  year={2000},
  publisher={Elsevier}
}

@book{turchetti2004stochastic,
  title={Stochastic models of neural networks},
  author={Turchetti, Claudio},
  volume={102},
  year={2004},
  publisher={IOS Press}
}

@article{costarelli2014convergence,
  title={Convergence of a family of neural network operators of the Kantorovich type},
  author={Costarelli, Danilo and Spigler, Renato},
  journal={J. Approx. Theory},
  volume={185},
  pages={80--90},
  year={2014},
  publisher={Elsevier}
}

@article{costarelli2022quantitative-esti.Normalized-and-kantoro.,
  title={Quantitative estimates for neural network operators implied by the asymptotic behaviour of the sigmoidal activation functions},
  author={Coroianu, Lucian and Costarelli, Danilo and Kadak, U{\u{g}}ur},
  journal={Mediterr. J. Math.},
  volume={19},
  number={5},
  pages={211},
  year={2022},
  publisher={Springer}
}

@article{anastassiou2022brownian-app.byNNO,
  title={Brownian motion approximation by neural networks},
  author={Anastassiou, George A and Kouloumpou, Dimitra},
  journal={Commun. Optim. theory},
  year={2022}
}

@article{anastassiou2023neural-Time-sepratingNNO,
  title={Neural Network Approximation for Time Splitting Random Functions},
  author={Anastassiou, George A and Kouloumpou, Dimitra},
  journal={Mathematics},
  volume={11},
  number={9},
  pages={2183},
  year={2023},
  publisher={MDPI}
}

@article{anastassiou2023approximation-TIMEsepatingNNOs-MDPI,
  title={Approximation of Brownian Motion on Simple Graphs},
  author={Anastassiou, George A and Kouloumpou, Dimitra},
  journal={Mathematics},
  volume={11},
  number={20},
  pages={4329},
  year={2023},
  publisher={MDPI}
}

@article{anastassiou2024brownian-timeseprating,
  title={Brownian motion approximation by parametrized and deformed neural networks},
  author={Anastassiou, George A and Kouloumpou, Dimitra},
  journal={Rev. R. Acad. Cienc. Exactas Fís. Nat. Ser. A Mat.},
  volume={118},
  number={1},
  pages={14},
  year={2024},
  publisher={Springer}
}

@article{fabiani2025random,
  title={Random projection neural networks of best approximation: Convergence theory and practical applications},
  author={Fabiani, Gianluca},
  journal={SIAM J. Math. Data Sci.},
  volume={7},
  number={2},
  pages={385--409},
  year={2025},
  publisher={SIAM}
}

@article{bolcskei2019optimal,
  title={Optimal approximation with sparsely connected deep neural networks},
  author={Bolcskei, Helmut and Grohs, Philipp and Kutyniok, Gitta and Petersen, Philipp},
  journal={SIAM J. Math. Data Sci.},
  volume={1},
  number={1},
  pages={8--45},
  year={2019},
  publisher={SIAM}
}

@article{yang2025rates,
  title={On the rates of convergence for learning with convolutional neural networks},
  author={Yang, Yunfei and Feng, Han and Zhou, Ding-Xuan},
  journal={SIAM J. Math. Data Sci.},
  volume={7},
  number={4},
  pages={1755--1772},
  year={2025},
  publisher={SIAM}
}

@article{archibald2024numerical,
  title={Numerical analysis for convergence of a sample-wise backpropagation method for training stochastic neural networks},
  author={Archibald, Richard and Bao, Feng and Cao, Yanzhao and Sun, Hui},
  journal={SIAM J. Numer. Anal.},
  volume={62},
  number={2},
  pages={593--621},
  year={2024},
  publisher={SIAM}
}

@article{isaacson2022mean,
  title={Mean field limits of particle-based stochastic reaction-diffusion models},
  author={Isaacson, Samuel A and Ma, Jingwei and Spiliopoulos, Konstantinos},
  journal={SIAM J. Math. Anal.},
  volume={54},
  number={1},
  pages={453--511},
  year={2022},
  publisher={SIAM}
}

@article{newman2022slimtrain,
  title={slimTrain---A Stochastic Approximation Method for Training Separable Deep Neural Networks},
  author={Newman, Elizabeth and Chung, Julianne and Chung, Matthias and Ruthotto, Lars},
  journal={SIAM J. Sci. Comput.},
  volume={44},
  number={4},
  pages={A2322--A2348},
  year={2022},
  publisher={SIAM}
}

@article{inglis2015mean,
  title={Mean-field limit of a stochastic particle system smoothly interacting through threshold hitting-times and applications to neural networks with dendritic component},
  author={Inglis, James and Talay, Denis},
  journal={SIAM J. Math. Anal.},
  volume={47},
  number={5},
  pages={3884--3916},
  year={2015},
  publisher={SIAM}
}

@article{costarelli2024asymptotic,
  title={Asymptotic analysis of neural network operators employing the Hardy-Littlewood maximal inequality},
  author={Costarelli, Danilo and Piconi, Michele},
  journal={Mediterr. J. Math.},
  volume={21},
  number={7},
  pages={199},
  year={2024},
  publisher={Springer}
}

@article{saini2025constructive,
  title={Constructive Approximation of Random Process via Stochastic Interpolation Neural Network Operators},
  author={Saini, Sachin and Singh, Uaday},
  journal={arXiv preprint arXiv:2512.24106},
  year={2025}
}
\end{document}